\documentclass[sigconf]{acmart}

\AtBeginDocument{%
  }

\copyrightyear{2025}
\acmYear{2025}
\setcopyright{cc}
\setcctype{by}
\acmConference[MM '25]{Proceedings of the 33rd ACM International Conference on Multimedia}{October 27--31, 2025}{Dublin, Ireland}
\acmBooktitle{Proceedings of the 33rd ACM International Conference on Multimedia (MM '25), October 27--31, 2025, Dublin, Ireland}\acmDOI{10.1145/3746027.3758171}
\acmISBN{979-8-4007-2035-2/2025/10}

\usepackage{comment}
\usepackage{bbm}
\usepackage[table]{xcolor}
\usepackage{pifont}
\usepackage{gensymb}

\begin{document}

\title{Pre-Forgettable Models: Prompt Learning as a Native Mechanism for Unlearning}


%

\author{Rutger Hendrix}
\affiliation{%
  \institution{University of Catania}
  \city{Catania}
  \country{Italy}}
\email{rutger.hendrix@phd.unict.it}

\author{Giovanni Patan\`e}
\affiliation{%
  \institution{University of Catania}
  \city{Catania}
  \country{Italy}}
\email{patane.giovanni@phd.unict.it}

\author{Leonardo G. Russo}
\affiliation{%
  \institution{University of Catania}
  \city{Catania}
  \country{Italy}}
\email{leonardo.russo@studium.unict.it}

\author{Simone Carnemolla}
\affiliation{%
  \institution{University of Catania}
  \city{Catania}
  \country{Italy}}
\email{simone.carnemolla@phd.unict.it}

\author{Federica Proietto Salanitri}
\affiliation{%
  \institution{University of Catania}
  \city{Catania}
  \country{Italy}}
\email{federica.proiettosalanitri@unict.it}

\author{Giovanni Bellitto}
\affiliation{%
  \institution{University of Catania}
  \city{Catania}
  \country{Italy}}
\email{giovanni.bellitto@unict.it}

\author{Concetto Spampinato}
\affiliation{%
  \institution{University of Catania}
  \city{Catania}
  \country{Italy}}
\email{concetto.spampinato@unict.it}

\author{Matteo Pennisi}
\affiliation{%
  \institution{University of Catania}
  \city{Catania}
  \country{Italy}}
\email{matteo.pennisi@unict.it}

\renewcommand{\shortauthors}{Rutger Hendrix et al.}

\begin{abstract}

Foundation models have transformed multimedia analysis by enabling robust and transferable representations across diverse modalities and tasks. However, their static deployment conflicts with growing societal and regulatory demands—particularly the need to unlearn specific data upon request, as mandated by privacy frameworks such as the GDPR. Traditional unlearning approaches, including retraining, activation editing, or distillation, are often computationally expensive, fragile, and ill-suited for real-time or continuously evolving systems.
In this paper, we propose a paradigm shift: rethinking unlearning not as a retroactive intervention but as a \textit{built-in capability}. We introduce a prompt-based learning framework that unifies knowledge acquisition and removal within a single training phase. Rather than encoding information in model weights, our approach binds class-level semantics to dedicated prompt tokens. This design enables instant unlearning simply by removing the corresponding prompt—without retraining, model modification, or access to original data.

Experiments demonstrate that our framework preserves predictive performance on retained classes while effectively erasing forgotten ones. Beyond utility, our method exhibits strong privacy and security guarantees: it is resistant to membership inference attacks, and prompt removal prevents any residual knowledge extraction, even under adversarial conditions. This ensures compliance with data protection principles and safeguards against unauthorized access to forgotten information, making the framework suitable for deployment in sensitive and regulated environments.
Overall, by embedding removability into the architecture itself, this work establishes a new foundation for designing modular, scalable and ethically responsive AI models.

\end{abstract}

\ccsdesc[500]{Computing methodologies~Learning settings}

\keywords{Machine unlearning, Parameter-efficient finetuning.}


\maketitle
\section{Introduction}
The advent of foundation models has significantly advanced multimedia understanding by enabling robust and transferable representations across a wide range of modalities and tasks~\cite{dosovitskiy2020image, ramesh2021zero, he2022masked}. These models, often pretrained on large-scale visual, textual, or multimodal datasets, form the backbone of state-of-the-art systems in image captioning, video retrieval, speech analysis, and more. However, despite their utility, the static and opaque nature of foundation models poses a growing concern in privacy-sensitive, safety-critical, and dynamically evolving domains. In such settings, models are not only expected to learn from large volumes of data but also to unlearn—that is, to remove specific knowledge or data influence efficiently, verifiably, and on demand.

Unlearning refers to the process of removing the contribution of certain data from a trained model while preserving its utility on the rest of the distribution. This is increasingly crucial in multimedia applications that involve personal, biometric, or copyrighted content. For example, removing user data from recommender systems, retracting learned associations from mistaken annotations in vision-language models, or erasing sensitive speech recordings from audio models all call for effective unlearning capabilities. Regulatory frameworks such as the GDPR~\cite{voigt2017eu} institutionalize this demand through the “right to be forgotten,” mandating systems to eliminate data traces upon request. Beyond compliance, unlearning supports fairness, adaptability, and continual learning, helping models evolve while avoiding the retention of outdated or biased knowledge~\cite{takshi2020unexpected}.

A recent survey by Nasirigerdeh et al.~\cite{unlearning_survey} categorizes existing techniques into three main families: exact unlearning (e.g., retraining from scratch), certified removal (e.g., influence function reversal), and approximate forgetting using parameter perturbations or distillation. While these methods are theoretically sound, most rely on post hoc interventions that require retraining, auxiliary models, or access to the original training data—factors that hinder deployment in dynamic, privacy-sensitive, or real-time systems.

Several more targeted strategies aim to reduce this overhead. For instance, SISA~\cite{bourtoule2021machine} partitions training data for selective retraining; gradient-negation techniques~\cite{mahadevan2022certifiable} attempt to counteract specific representations; and decomposition-based methods~\cite{kodge2024deep} isolate class-discriminative features in latent space. Others, such as SSD~\cite{foster2024fast} and EU-k/CF-k~\cite{goel2022towards}, modify only subsets of parameters using importance-based reweighting or layer-wise resets. Teacher-student models~\cite{chundawat2023can} and prompt-based strategies~\cite{xu2024lmeraser} offer more modular mechanisms, but still require separate unlearning stages and incur nontrivial training costs. In practice, most of these methods suffer from one or more critical limitations: the need for extensive retraining, assumptions about architecture or data availability, and limited compatibility with frozen foundation models. These constraints make them difficult to adopt in real-world multimedia systems, where models must remain continuously available, adaptable, and compliant with evolving privacy requirements.

To address these limitations: embedding unlearning directly into the learning process, rather than relying on post hoc corrections. Our goal is to design a mechanism that enables modular, efficient, and interpretable unlearning, aligned with the demands of real-world deployment. At a high level, our strategy leverages the emerging paradigm of prompt-based learning, but reinterprets its purpose. While prior work~\cite{lester2021power, liu2021p, zhou2022learning} primarily uses prompts to enhance task adaptation, we treat them as containers for class-specific knowledge that can be independently manipulated, specifically, removed. This transforms prompts into modular units of representation, decoupled from the model’s core parameters. Concretely, we introduce a class-specific prompting framework in which each class is associated with a dedicated, learnable prompt token. During training, only the prompts, adapters, and classifier head are updated, while the foundation model backbone remains frozen. This ensures that knowledge remains localized within prompts, enabling immediate and targeted unlearning by simply removing the corresponding prompt—without modifying the model’s weights, accessing training data, or triggering any retraining procedure.

This design has several important implications. First, it enables a native mechanism (without requiring any retraining)  for unlearning: forgetting is achieved on-the-fly, even during sequential inference. Second, it supports data-free and interpretable unlearning—removal of a prompt leads to abstention or uniform predictions, clearly indicating the absence of learned knowledge. Finally, the frozen backbone prevents interference with retained classes and ensures computational efficiency and robustness, even in continual learning or edge settings. Collectively, these properties establish our framework as a practical and scalable solution for unlearning in foundation models.

To validate our approach, we conducted extensive experiments across three medical image classification tasks from the MedMNIST v2 benchmark—BloodMNIST, DermaMNIST, and OrganMNIST—using a frozen Vision Transformer (ViT) with shallow prompt tuning and LoRa adapters. We evaluated unlearning by measuring performance on retained and forgotten classes under three settings: forgetting one class, half of the classes, or all but one. Additional experiments were performed on UrbanSound8K (audio) to assess applicability across modalities. Across all datasets, our method achieves near-random accuracy on forgotten classes while maintaining high performance on retained ones—without retraining, architectural modification, or access to original data. Ablation and privacy analyses further confirm the method’s efficiency, interpretability, and robustness. These results highlight the key advantages of our framework: it is modular, retraining-free, data-free, and inherently privacy-preserving—making it uniquely suited for dynamic, regulation-compliant deployments.

In summary, we introduce a new class of adaptive models that embed unlearning into their design from the outset. Prompt-based optimization emerges as a powerful mechanism for enabling privacy-preserving, efficient, and interpretable unlearning—shifting the paradigm from reactive post hoc solutions to proactive, pre-forgettable architectures.
\section{Related Work}
Machine unlearning has become an increasingly important area of study, driven by privacy regulations such as the GDPR and the growing demand for dynamic data governance in AI systems. Existing techniques can be broadly classified into two categories: data-centric and model-centric approaches, each presenting distinct trade-offs in terms of scalability, guarantees, and practicality.

\textbf{Data-centric} methods aim to localize data influence within the training set to simplify removal operations. For instance, SISA~\cite{bourtoule2021machine} partitions the data into disjoint shards and trains separate sub-models, allowing efficient retraining upon deletion. While this avoids full model retraining, it requires storage overhead and may degrade performance as the number of shards increases. Other strategies include data obfuscation~\cite{graves2021amnesiac}, which corrupts or relabels targeted samples, and summation-based representations~\cite{cao2015towards}, which compress training data to support fast recomputation. However, these methods typically assume access to the original dataset and offer limited guarantees of complete forgetting.

\textbf{Model-centric} techniques instead modify the trained model directly, without relying on the training data. Certified removal~\cite{guo2020certified} uses influence functions to estimate and reverse the effect of individual data points—mainly in convex settings. SSD~\cite{foster2024fast}, a Fisher-based approach, perturbs parameters according to their importance to selectively erase memorized information. While these methods bypass retraining and offer some theoretical grounding, they often incur high computational costs and face challenges in deep, non-convex architectures. Hybrid techniques like CF-$k$ and EU-$k$~\cite{goel2022towards} achieve partial unlearning by retraining upper layers while freezing lower ones, but this can leave residual knowledge in the frozen backbone, limiting their effectiveness.

Building on these categories, \textbf{class-level unlearning} has emerged as a more structured and task-aligned formulation, especially relevant for classification problems. Instead of targeting individual samples, these methods aim to erase entire semantic categories. PBU~\cite{panda2024partially} treats unlearning as a Bayesian optimization process, selectively degrading performance on the forgotten class while maintaining accuracy elsewhere. More recently, prompt-based techniques~\cite{xu2024lmeraser} associate clustered data regions with learned prompts and classifier heads, enabling post hoc removal through retraining specific clusters. While promising, these methods still require additional unlearning phases and lack true modularity or data-free operation.

Our method extends this line of work by reimagining prompts as self-contained carriers of class-specific knowledge, explicitly localized during training. By binding each class to a dedicated prompt token and freezing the backbone, our model enables immediate, verifiable unlearning simply by removing the corresponding prompt—without accessing data, modifying model weights, or performing any retraining. This design supports scalable, interpretable, and retraining-free unlearning, directly addressing the limitations of prior approaches in both practicality and deployability.

\section{Method}
We propose a prompt-based framework that attaches a \emph{learnable prompt} to each class in order to gate its recognition. In essence, the knowledge of each class is isolated into a small prompt vector (or set of vectors) that conditions the model’s prediction for that class. This design is modality-agnostic: it can be applied to image classification, text classification, or any domain where a pretrained encoder can be conditioned on prompts. Crucially, \textbf{unlearning a class is as simple as removing its prompt}—with no retraining and no changes to the encoder’s weights. Once a class’s prompt is removed, the model no longer recognizes that class, as if the class had never been learned, while all other classes remain fully recognizable.

For example, consider a medical image classification scenario in which a model diagnoses diseases from X-ray images. In our framework, each disease class (e.g., pneumonia, nodule, etc.) is associated with its own prompt embedding that “activates” recognition of that condition. During inference, an X-ray image is analyzed in conjunction with each disease’s prompt to produce class-specific confidence scores. If a certain disease must be unlearned (say due to a patient data removal request or a shift in policy), we simply detach the corresponding disease prompt. The model, without that prompt, will cease to predict the removed disease—any input that previously triggered the now-removed prompt will yield no high-confidence prediction for that class. This medical imaging use case illustrates the approach in a high-stakes domain, but the method is general: the prompt-based gating works for any classification task (but it can be extended to all those tasks that can be tackled through prompt tuning), providing a flexible mechanism to add or remove classes on the fly.

\begin{figure*}[htb!]
    \centering
    \includegraphics[width=0.9\textwidth]{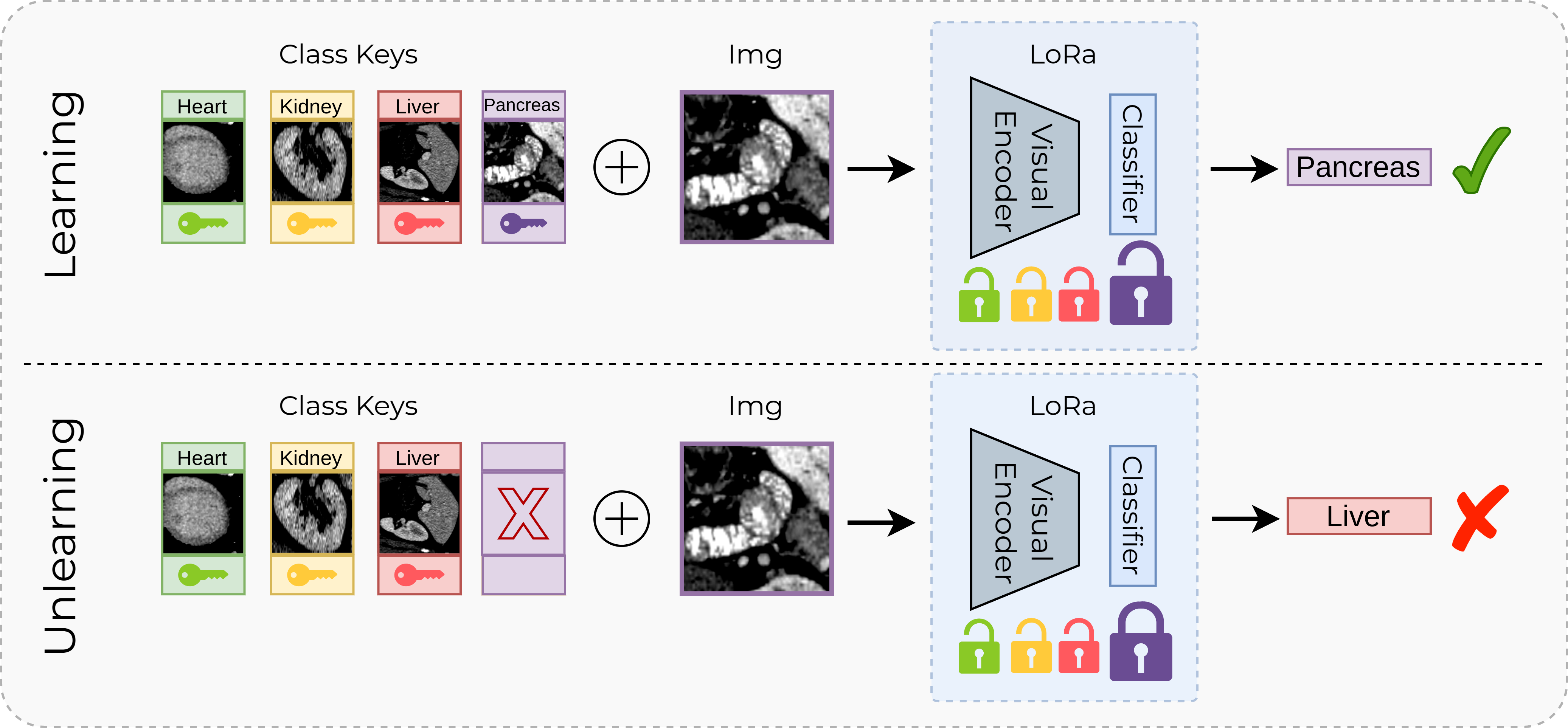}
    \caption{\textbf{Prompt-based learning and unlearning framework for medical image classification.} Each class is linked to a specific prompt. During training, prompts are used to learn class-specific representations. At inference, removing a prompt prevents the model from recognizing that class.}
    \label{fig:method}
\end{figure*}

Figure~\ref{fig:method} illustrates the mechanism using a medical image classification case: during training, the model learns to associate each class with its unique prompt (key). During inference, removing the prompt prevents recognition of that class, effectively achieving unlearning.

The architecture of the proposed approach consists of a frozen backbone encoder $f_{\theta}$ (with parameters $\theta$ fixed after pre-training) and a set of class-specific prompt embeddings. Each class $c$ in the label set $C$ is assigned a prompt $p_c$, which is a small learnable vector or collection of vectors. These prompts serve as {\it dynamic conditioning parameters} that modulate the encoder’s behavior for their respective class. We represent each prompt $p_c$ as, for example, a trainable sequence of $M$ tokens in the same embedding dimension $d$ as the encoder. 

The prompts are integrated into the input processing pipeline of the encoder through concatenate to the input features: for vision transformers (e.g., ViT backbone used for medical image classification), we prepend the $M$ prompt tokens to the sequence of image patch embeddings at the input layer. This way, the self-attention layers will attend to both the image patches and the prompt tokens. 
. 

\subsection{Formalisation}

Let \( f_\theta \) be a frozen pretrained foundation model suitable for the task domain, e.g., a vision transformer (ViT) for image classification or a CLIP-like encoder for multimodal inputs—with fixed parameters \( \theta \). The model is never updated during training, which ensures that learned representations remain stable and that changes to prompt structure are fully traceable.

We define a set of target classes \( \mathcal{C} = \{c_1, \dots, c_K\} \). To each class \( c_i \in \mathcal{C} \), we associate a unique learnable prompt vector \( p_{c_i} \). These prompt vectors are the only trainable components of our system and collectively form the prompt set \( P = \{p_{c_1}, p_{c_2}, \dots, p_{c_K}\} \). Each prompt is intended to function as a semantic key that unlocks the model's ability to recognize the corresponding class.

Let \( x \in \mathcal{X} \) denote an input sample and \( y \in \mathcal{Y} \) its corresponding label. For each training instance, we provide the model with the correct class prompt \( p_c \) along with a set of \( m \) distractor prompts, sampled from the remaining class prompts \( P \setminus \{p_c\} \). This forces the model to learn to discriminate based on the correct prompt, even in the presence of misleading alternatives.

The prediction is thus made by the following formulation:
\begin{equation}
\hat{y} = f_\theta(x, \mathcal{S}(\{p_c\} \cup \sigma(P \setminus \{p_c\}, m)))
\end{equation}
where \( \mathcal{S} \) is a permutation function used to randomize prompt order and prevent positional bias, and \( \sigma(P \setminus \{p_c\}, m) \) is a sampling function defined below.

To construct the distractor prompt set, we define the sampling operation:
\begin{equation}
\begin{split}
\sigma(P \setminus \{p_c\}, m) = \{p_{c_1}, p_{c_2}, \dots, p_{c_m}\}, \\
\text{where } \{p_{c_1}, \dots, p_{c_m}\} \sim \text{Uniform}(\mathcal{P}_m(P \setminus \{p_c\}))
\end{split}
\end{equation}
This ensures that the model is consistently exposed to a diverse mixture of incorrect class prompts during training, which strengthens the coupling between correct prompt and class label, while weakening associations with unrelated prompts.

Thanks to this design, forgetting a class is immediate and straightforward: to unlearn class $k$, we remove its prompt $p_k$ from the set $P$. This removal means that for any input $x$, the posterior $P(c \mid x, P \setminus \{p_k\})$ will be re-normalized over the remaining classes $C \setminus \{k\}$. If the model has been trained such that class $k$’s features do not inadvertently activate other prompts, then after removing $p_k$ the model will abstain from predicting $k$ and typically yield either a low-confidence prediction among the rest or a nearly uniform output distribution for inputs of class $k$. 

\subsection{Training objective}

The training objective is designed to simultaneously achieve two complementary goals: (i) accurate classification when the correct class prompt is present, and (ii) effective unlearning when the class prompt is absent. To this end, we define two loss terms—a standard learning objective and a novel unlearning objective—that are jointly optimized.

\paragraph{Learning Objective.} The primary objective ensures that the model correctly predicts the class label \( y \) when the correct prompt \( p_c \) is available among the input prompts. This encourages the model to associate the presence of a specific prompt with the corresponding class semantics. The learning loss is computed as:

\begin{equation}
\mathcal{L}_{\text{learn}} = \mathbb{E}_{(x,c) \sim \mathcal{D}} \left[ \ell \left( f_{\theta}(x, \mathcal{S}(\{p_c\} \cup \sigma(P \setminus \{p_c\}, m))), y \right) \right]
\end{equation}

where \( \ell \) is the standard cross-entropy loss, \( \mathcal{S} \) denotes the permutation operator applied to the set of prompts (to prevent position encoding bias), and \( \sigma \) is the distractor sampling function described earlier. This loss enforces that classification performance remains high under the correct prompt-conditioning scenario.

\paragraph{Unlearning Objective.} To ensure that the model abstains from predicting a class when its corresponding prompt is removed, we introduce an unlearning objective. Specifically, when the input is paired only with incorrect prompts (i.e., the target class prompt is excluded), the model is trained to produce a high-entropy output distribution over the class space. This is achieved by minimizing the divergence between the model’s output and a uniform distribution:

\begin{equation}\label{eq:unlearning}
\mathcal{L}_{\text{unlearn}} = \mathbb{E}_{(x, c) \sim \mathcal{D}, m \sim p(m)} \left[ D_{\text{KL}} \left( f_{\theta}(x, \mathcal{S}(\sigma(P \setminus \{p_c\}, m))) \Big\| \mathbf{u} \right) \right]
\end{equation}

where \( \mathbf{u} \) is the uniform distribution over the \( K \) classes, and \( D_{\text{KL}} \) denotes the Kullback-Leibler divergence. This term penalizes confident predictions made in the absence of the correct prompt, promoting uncertainty and thus enabling effective unlearning behavior.

The final training loss combines the two components described above:

\begin{equation}
\mathcal{L} = \mathcal{L}_{\text{learn}} + \lambda \mathcal{L}_{\text{unlearn}}
\end{equation}

where \( \lambda \in \mathbb{R}^+ \) is a hyperparameter controlling the trade-off between accurate classification and the strength of the unlearning signal.

\medskip
\subsection{Unlearning-Driven Design Advantages}
This unified optimization scheme ensures that each prompt functions as both a semantic key and a control token: the model learns to predict a class only when its associated key is present and reliably abstains when it is absent. By encoding this behavior during training, we eliminate the need for post hoc unlearning procedures, achieving fast, interpretable, and modular forgetting directly at inference time. This makes the approach particularly appealing for dynamic or privacy-sensitive multimedia systems, where class availability may evolve over time or depend on external constraints (e.g., regulatory requests, user consent, or clinical scope).

Overall, the advantages of the proposed formulation are the following ones:
\begin{itemize}
\item \textbf{Modularity.} Our framework is highly modular: each class is encapsulated by its prompt, acting as an independent module that can be added or removed at will. Eliminating a class is done by simply dropping its prompt. The rest of the system continues to function with zero change.

\item \textbf{Interpretability.} The effect of removing knowledge is transparent in this framework. When a prompt is removed, the model explicitly loses the ability to recognize the associated class, which is reflected in its outputs: if an image of the removed class is given, no prompt will trigger a high response, resulting in either an abstention (no class scores high) or a roughly uniform probability across remaining classes. This behavior is intuitively interpretable as “the model doesn’t know that class.” In contrast to opaque weight pruning or obscure fine-tuning adjustments, our method provides a direct correspondence between a unit (the prompt) and a semantic concept (the class). Thus we can understand and control the model’s knowledge at the granularity of classes.

\item \textbf{Scalability.} Because the backbone encoder is fixed, scaling the model to many classes is efficient. The memory and computational footprint of prompts is minimal – prompt vectors are tiny compared to model weights – so adding dozens of new classes only increases the model size and inference cost marginally. There is no need to retrain or even touch the heavy backbone $\theta$ when the class set changes, avoiding the expensive retraining typically required for updating a classifier. 

\end{itemize}

\section{Experiments}

\newcommand{\result}[2]{$\mathbf{#1} \pm #2$} 

We evaluate our method through a comprehensive set of experiments designed to assess its effectiveness, robustness, and generalizability in achieving class-level unlearning. Given the increasing demand for privacy-aware machine learning—especially in sensitive domains like healthcare—our primary evaluation focuses on medical image classification, where forgetting disease-specific classes is of practical importance. To demonstrate the modality-agnostic nature of our approach, we extend the evaluation to audio classification. The evaluation protocol is structured to quantify not only how well the model forgets specified classes but also how effectively it retains performance on the remaining ones. In addition, we assess privacy resilience and computational efficiency, thereby establishing the practicality of our method in real-world dynamic learning environments.

\subsection{Datasets}

\noindent For medical image classification, we used three publicly available medical image classification datasets as part of the MedMNIST v2 collection \cite{medmnistv2}: BloodMNIST, DermaMNIST, and OrganSMNIST. These datasets span diverse medical imaging modalities and diagnostic tasks, allowing us to assess the generalizability of our method across different domains:
	\begin{itemize}
	    \item \textit{BloodMNIST} \cite{bloodmnist}: Comprises 17,092 microscopy images of blood cells annotated for classification into eight categories, including different types of white blood cells, red blood cells, and platelets. 
\item \textit{DermaMNIST} \cite{dermamnist1, dermamnist2}: A dataset of 10,015 dermatoscopic images representing seven common pigmented skin lesions.
	   \item \textit{OrganSMNIST} \cite{organmnist1}: Contains 25,211 abdominal CT scans for organ classification, covering 11 different organ types. 
\end{itemize}
All images have a resolution 224 × 224 pixels, and we follow the standard dataset splits for training, validation, and testing, as described in the medMNIST v2 collection.

For the audio classification task, we use the UrbanSound8K \cite{salamon2014dataset} a widely adopted benchmark for environmental sound classification, consisting of 8,732 labeled audio clips, each up to 4 seconds in duration, covering 10 distinct sound categories. 

\subsection{Experimental setup}

For medical image classification, we use a Vision Transformer (ViT-Base) as the frozen backbone and adapt it through parameter-efficient fine-tuning. For audio classification, we, instead, use the Audio Spectrogram Transformer (AST) backbone \cite{gong2021ast}, a convolution-free Transformer model pre-trained on large-scale audio datasets. In all the two scenarios, only the classifier head, prompts, and adapters are trained while keeping the backbone fixed. Visual Prompt Tuning (VPT)~\cite{jia2022visual} is applied in a shallow configuration. LoRa adapters ~\cite{hu2022lora} are integrated into each Query and Value matrix of the transformer layers with a low bottleneck dimension  $r=4$  and a scaling factor $s=4$. A learning rate of  $1e^{-3}$  is used for all learnable parameters. 
The model is trained for 10 epochs following the approach described in Section 3, where class-specific prompts are optimized simultaneously for learning and unlearning. 
We also evaluate performance of a \textit{Full Knowledge} model, which follows the same configuration as above, but is trained solely for learning using a standard cross-entropy loss (i.e., we do not apply the unlearning loss defined in Formula~\ref{eq:unlearning}. This provides an upper bound for accuracy retention after unlearning.  

To assess the effectiveness of unlearning, we measure accuracy on two sets: the retain set and the forget set. The \textbf{retain accuracy}, $Acc_r$, computed as

\begin{equation}
    \text{Acc}_r = \frac{1}{|\mathcal{C}_r|} \sum_{c \in \mathcal{C}_r} \mathbb{E}_{(x, c) \sim \mathcal{D}_r} \mathbbm{1} \left[ f_{\theta}(x) = c \right]
\end{equation}

quantifies the classification performance on the classes that should be retained, where \( \mathcal{C}_r \) represents the set of retained classes, \( \mathcal{D}_r \) is the corresponding data distribution and $\mathbbm{1}$ is the indicator function. 

The \textbf{forget accuracy}, $Acc_f$, computed as

\begin{equation}
    \text{Acc}_f = \frac{1}{|\mathcal{C}_f|} \sum_{c \in \mathcal{C}_f} \mathbb{E}_{(x, c) \sim \mathcal{D}_f} \mathbbm{1} \left[ f_{\theta}(x) = c \right]
\end{equation}

measures how well the model still classifies forgotten classes, where \( \mathcal{C}_f \) represents the set of forgotten classes and \( \mathcal{D}_f \) is the corresponding data distribution.
An effective unlearning method should reduce forget accuracy ($\text{Acc}_f$) to near-random levels while maintaining stable retain accuracy ($\text{Acc}_r$), ensuring that forgotten knowledge is erased without affecting retained classes.  

We consider three incremental scenarios based on the number $f$ of classes to be forgotten. In the first scenario, \( f = 1 \), only a single class is removed, leaving all others intact.  In the second scenario, half of the classes are removed, such that \( f = C / 2 \)  where \( C \) is the total number of classes in the dataset. In the third scenario, all but one class are forgotten, which means \( f = C - 1 \). Per scenario, we report the average of all possible combinations.

\begin{table*}[ht]
\centering
\setlength{\tabcolsep}{2pt}
\caption{Unlearning results for $f = 1$, where a single class is forgotten. Notably, all competing methods require 10 additional epochs of unlearning, whereas our method achieves effective forgetting without any retraining—demonstrating its retraining-free capability and deployment efficiency.}

\begin{tabular}{c|cc|cc|cc}
\toprule
\textbf{Method }& \multicolumn{2}{c|}{\textbf{BloodMNIST}}                        & \multicolumn{2}{c|}{\textbf{DermaMNIST}}  & \multicolumn{2}{c}{\textbf{OrganSMNIST}}\\
                     \midrule
\textbf{Full Knowledge} & \multicolumn{2}{c|}{98.60}  & \multicolumn{2}{c|}{85.99} & \multicolumn{2}{c}{79.77} \\
\midrule
 & $\text{Acc}_r$     &          $\text{Acc}_f$  &$\text{Acc}_r$  & $\text{Acc}_f$ &$\text{Acc}_r$ & $\text{Acc}_f$   \\
\midrule
\rowcolor{gray!10}
\textbf{ SSD~\cite{foster2024fast}}   & \result{94.06}{02.03}  & \result{07.68}{01.12}       & \result{76.47}{07.30} & \result{13.40}{05.34}  & \result{71.65}{04.02}             & \result{07.14}{03.85}  \\
\textbf{ Goel et al.~\cite{goel2022towards}} &   \result{81.81}{01.90}  & \result{10.95}{01.75}       & \result{76.87}{01.75} & \result{43.83}{02.29}  & \result{76.86}{02.41}             & \result{44.33}{06.35}  \\
\rowcolor{gray!10}
\textbf{ LMEraser~\cite{xu2025lmeraser}} &   \result{91.71}{06.00}  & \result{0.02}{0.06}       & \result{79.67}{04.46} & \result{00.01}{0.03}  & \result{62.86}{10.03}             & \result{00.13}{00.23}  \\

\textbf{Ours} & \result{97.61}{02.63}  & \result{10.54}{11.38}       & \result{85.14}{04.55}  & \result{18.84}{09.27}    & \result{82.19}{02.63}              & \result{07.69}{08.47} \\
\bottomrule
\end{tabular}
\label{tab:results_1}
\end{table*}

\begin{table*}[ht]
\centering
\setlength{\tabcolsep}{2pt}
\caption{\textbf{Unlearning results for $ f = C/2 $, where half of the classes are forgotten. Competitors are re-trained for 10 epochs.} }
\begin{tabular}{c|cc|cc|cc}
\toprule
\textbf{Method} &  \multicolumn{2}{c|}{\textbf{BloodMNIST}}                        & \multicolumn{2}{c|}{\textbf{DermaMNIST}}  & \multicolumn{2}{c}{\textbf{OrganSMNIST}}\\

                      \midrule
\textbf{Full Knowledge } & \multicolumn{2}{c|}{98.60}  & \multicolumn{2}{c|}{85.99} & \multicolumn{2}{c}{79.77} \\
\midrule
& $\text{Acc}_r$     &          $\text{Acc}_f$  &$\text{Acc}_r$  & $\text{Acc}_f$ &$\text{Acc}_r$ & $\text{Acc}_f$   \\
\midrule

\rowcolor{gray!10}
\textbf{SSD~\cite{foster2024fast}}    & \result{95.45}{01.51}   & \result{95.45}{03.01}        & \result{70.82}{00.81}  & \result{67.01}{01.17} &    \result{69.52}{04.32}  & \result{69.12}{04.50} \\
\textbf{Goel et al.~\cite{goel2022towards}}  & \result{95.56}{08.30}  & \result{95.56}{10.40}       & \result{70.87}{07.30} & \result{67.06}{08.15} & \result{79.83}{11.27}  & \result{62.58}{10.06} \\
\rowcolor{gray!10}
\textbf{LMEraser~\cite{xu2025lmeraser}}  & \result{97.75}{01.38}  & \result{00.02}{00.04}       & \result{84.76}{06.62} & \result{00.03}{00.04}  & \result{81.41}{07.56}  & \result{00.01}{00.00} \\
\textbf{Ours}   & \result{98.43}{01.37}    & \result{12.74}{08.86}        & \result{87.52}{05.68}   & \result{16.00}{05.86}    & \result{86.62}{05.93}             & \result{07.72}{04.52}  \\
\bottomrule
\end{tabular}
\label{tab:results_n2}
\end{table*}

\begin{table*}[ht]
\centering
\setlength{\tabcolsep}{2pt}
\caption{\textbf{Unlearning results for $ f = C - 1 $,} where all but one class is forgotten. Competitors are re-trained for 10 epochs.}

\begin{tabular}{c|cc|cc|cc}
\toprule
\textbf{Method} &  \multicolumn{2}{c|}{\textbf{BloodMNIST}}                        & \multicolumn{2}{c|}{\textbf{DermaMNIST}}  & \multicolumn{2}{c}{\textbf{OrganSMNIST}}\\

                      \midrule
\textbf{Full Knowledge } & \multicolumn{2}{c|}{98.60}  & \multicolumn{2}{c|}{85.99} & \multicolumn{2}{c}{79.77} \\
\midrule
& $\text{Acc}_r$     &          $\text{Acc}_f$  &$\text{Acc}_r$  & $\text{Acc}_f$ &$\text{Acc}_r$ & $\text{Acc}_f$   \\
\midrule

\rowcolor{gray!10}
\textbf{SSD~\cite{foster2024fast}}    & \result{73.29}{06.58}  & \result{77.36}{07.15}       & \result{56.36}{08.87} & \result{77.31}{06.33}  &    \result{66.30}{03.59}  & \result{71.72}{2.46}  \\
\textbf{Goel et al.~\cite{goel2022towards}}  & \result{77.39}{04.81}  & \result{68.80}{05.30}       & \result{59.45}{06.69} & \result{74.61}{08.39}  & \result{80.86}{03.93}  & \result{68.17}{04.89} \\ 
\rowcolor{gray!10}
\textbf{LMEraser~\cite{xu2025lmeraser}}  & \result{99.75}{00.31}  & \result{01.12}{01.73}       & \result{97.76}{02.33} & \result{00.35}{00.35}  & \result{99.80}{00.18}  & \result{01.61}{01.31} \\
\textbf{Ours}   & \result{99.27}{01.43}    & \result{16.22}{04.19}        & \result{96.09}{07.37}   & \result{19.82}{09.28}    & \result{98.51}{01.95}             & \result{11.80}{04.31}  \\
\bottomrule
\end{tabular}
\label{tab:results_n3}
\end{table*}

Among existing unlearning methods, we selected SSD ~\cite{foster2024fast} and Goel et al.~\cite{goel2022towards} and LMeraser ~\cite{xu2025lmeraser} due to their ease of adaptation across different datasets, and close resemblance to our instant and prompt unlearning objective. Methods requiring extensive modifications were excluded to prevent potential bias. To ensure a fair comparison, we performed a grid search for hyperparameter selection.
As a baseline model for the competitors, we trained the same ViT with LoRa and prompts for 10 epochs without unlearning loss, then fine-tuned it for 10 more epochs using their respective unlearning strategies.\\

\subsection{Unlearning performance}

Tables \ref{tab:results_1}, \ref{tab:results_n2}  and \ref{tab:results_n3} compare the performance of our method against SSD \cite{foster2024fast} and Goel et al. \cite{goel2022towards} and LMeraser \cite{xu2025lmeraser} on BloodMNIST, DermaMNIST, and OrganMNIST on forgetting a single class ($f = 1$), forgetting half of the classes ($f = C / 2$) and forgetting all but one class ($f = C - 1$) scenarios. 
In the first case ($f = 1$), our method significantly outperforms SSD and Goel et al. in terms of $\text{Acc}_r$, achieving the highest retain accuracy (in some cases performance even improves when unlearning) across all datasets, demonstrating that it does not degrade performance on the retained classes.  In terms of forgetting effectiveness, our method shows a moderate drop in $\text{Acc}_f$, with values of $10.54\%$ (BloodMNIST), $18.84\%$ (DermaMNIST) and $7.69\%$ (OrganSMNIST). Although SSD achieve lower $\text{Acc}_f$ on BloodMNIST and OrganSMNIST, it does so at the cost of significantly lower $\text{Acc}_r$, suggesting an excessive forgetting that negatively impacts retained classes. In contrast, Goel et al. exhibit weaker forgetting, with $\text{Acc}_f$ values above $40\%$ for DermaMNIST and OrganSMNIST, indicating incomplete unlearning. 

When half of the classes are forgotten ($f = C/2$), SSD and Goel et al. struggle to suppress forgotten information while preserving useful knowledge, failing to effectively reduce $\text{Acc}_f$ or maintain high $\text{Acc}_r$. Our methods consistently has chance-level accuracies for $\text{Acc}_f$, with values of $12.74\%$, $16.00\%$ and $7.72\%$, for BloodMNIST, DermaMNIST and OrganSMNIST, respectively.

\begin{figure*}[htb!]
    \centering
    \includegraphics[width=0.85\textwidth]{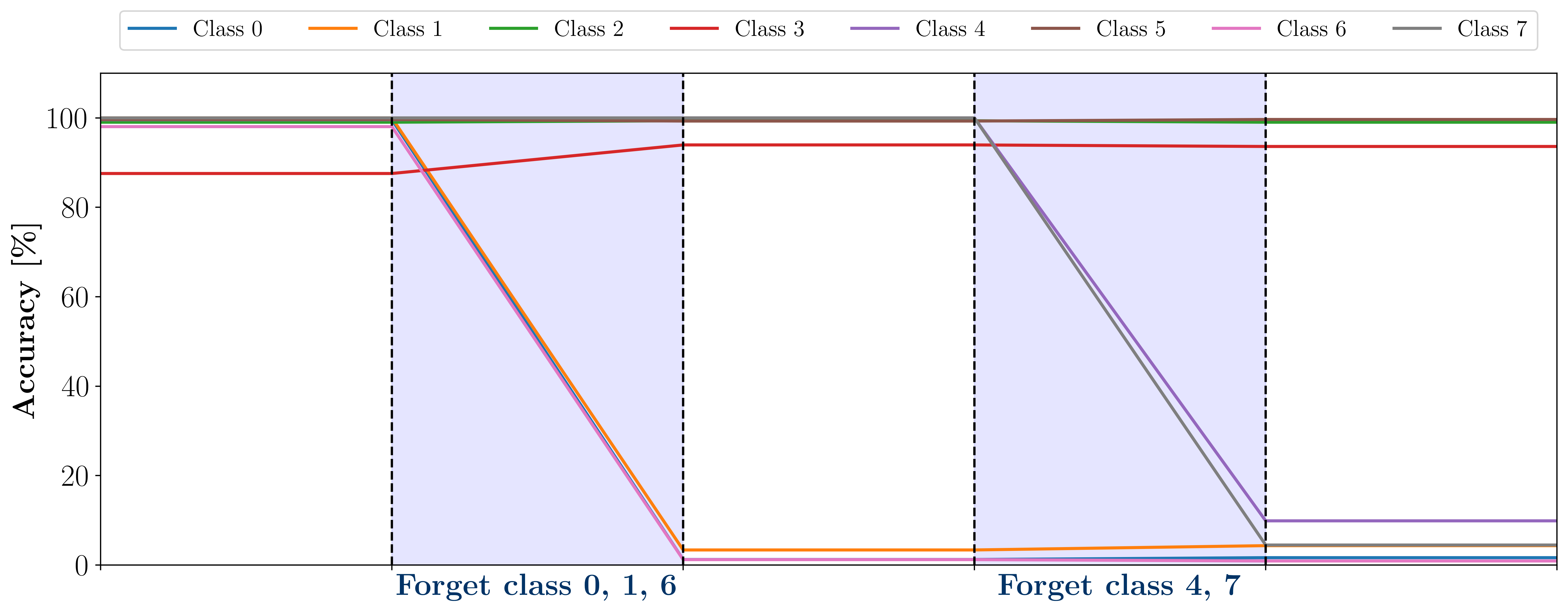}
    \caption{\textbf{Accuracy evolution during sequential inference}. Violet regions mark unlearning phases, where the corresponding classes (shown beneath) are forgotten by removing their prompts, causing a sharp drop in accuracy. Retained class accuracy, instead, remains stable throughout the entire inference phase.}
    \label{fig:acc_plot}
\end{figure*}

When nearly all classes are removed ($ f = C - 1 $), our method continues to show robust retention with $\text{Acc}_r$ values of $99.27\%$ on BloodMNIST, $96.09\%$ on DermaMNIST and $98.51\%$ on OrganSMNIST, maintaining a high classification performance on the single retained class. This is significantly higher than SSD and Goel et al., both of whom experience considerable drops in retain accuracy. Regarding forgetting, our approach reduces $\text{Acc}_f$ to $16.22\%$ on BloodMNIST, $19.82\%$ on DermaMNIST, and $11.80\%$ on OrganSMNIST, effectively suppressing information about forgotten classes. In contrast, SSD and Goel et al. exhibit significantly higher $\text{Acc}_f$, exceeding $70\%$, suggesting they fail to erase knowledge of forgotten classes.

Across all experimental settings, LMeraser achieves stronger unlearning performance (i.e., lower $\text{Acc}_f$) and comparable retain accuracy ($\text{Acc}_r$) relative to our method. However, this advantage comes at a substantial cost. LMeraser requires a dedicated post hoc unlearning phase involving 10 full epochs of retraining for each update, which significantly increases training time and memory overhead. Specifically, with 30 centroids, it retrains an average of 14.9, 22.4, and 28.7 clusters across the BloodMNIST, DermaMNIST, and OrganSMNIST datasets for $f = 1$, $f = C/2$, and $f = C{-}1$, respectively. As a consequence, the effective model size becomes 6.95× larger when forgetting a single class ($f = 1$) and grows up to 13.34× when forgetting all but one class ($f = C{-}1$).

In contrast, our method achieves effective unlearning with zero retraining—a key distinction that enables substantial gains in computational efficiency and deployment flexibility. Our retraining-free design supports on-the-fly forgetting during inference by simply removing the class-specific prompt tokens. As illustrated in Figure~\ref{fig:acc_plot}, this leads to an immediate and predictable drop in classification performance for forgotten classes, while preserving stable accuracy on retained ones.
Moreover, as the number of forgotten classes increases, the unlearning performance gap between LMeraser and our method progressively narrows, confirming the robustness and scalability of our strategy under larger deletion scenarios. This, combined with the absence of retraining, minimal memory overhead, and modular structure, positions our method as a better solution for real-time deployments than LMEraser.

We also performed an ablation study on BloodMNIST to evaluate the impact of KL divergence, shuffling $S(\cdot)$, and prompt sampling $\sigma(\cdot)$ on unlearning performance as the number $f$ of forgotten classes varies. When using only KL divergence, retain accuracy is significantly degraded ($\text{Acc}_r = 20.96\%$ for $ f = 3 $) while the forget accuracy remains high ($\text{Acc}_f = 24.10\%$ for $ f = 1 $), indicating that KL alone is insufficient for effective unlearning. Adding shuffling improves forgetting ($\text{Acc}_f = 16.04\%$ for $ f = 1 $) but still limits retention ($\text{Acc}_r = 49.25\%$ mean). The full method, incorporating prompt sampling, achieves the best trade-off, maintaining high retain accuracy while effectively reducing forget accuracy. 

\begin{table}[htb]
\centering
\setlength{\tabcolsep}{2pt}
\caption{\textbf{Ablation study }on the BloodMNIST dataset  as the number $f$ of forgotten classes varies}
\label{tab:results}
\begin{tabular}{ccc|cc|cc|cc|cc} 
\toprule
 & & & \multicolumn{2}{c|}{f=1} & \multicolumn{2}{c|}{f=3} & \multicolumn{2}{c|}{f=7} & \multicolumn{2}{c}{Mean}  \\
\cmidrule(lr){4-5} \cmidrule(lr){6-7} \cmidrule(lr){8-9} \cmidrule(lr){10-11}
KL & $S(\cdot)$  & $\sigma(\cdot)$  & $\text{Acc}_r$ & $\text{Acc}_f$ & $\text{Acc}_r$ & $\text{Acc}_f$ & $\text{Acc}_r$ & $\text{Acc}_f$   & $\text{Acc}_r$ & $\text{Acc}_f$  \\
\midrule
\rowcolor{gray!10}
\checkmark & \ding{55} & \ding{55}    & 71.79 & 24.10 & 20.96 & 18.03 & 23.43 & 17.72 & 38.73 & 19.95 \\
\checkmark & \checkmark & \ding{55}   & 66.73 & 16.04 & 40.67 & 21.16 & 40.37 & 25.38 & 49.25 & 20.86 \\
\rowcolor{gray!10}
\checkmark & \checkmark & \checkmark  & \textbf{97.61} & \textbf{10.54} & \textbf{98.4}\textbf{2} & \textbf{12.73} & \textbf{99.27} & \textbf{16.22} & \textbf{98.43} & \textbf{13.16} \\
\bottomrule
\end{tabular}
\end{table}

Extending our prompt unlearning strategy to other audio modality, Table~\ref{tab:results_audio} summarizes the performance of our method on an audio classification task. Across various configurations, the method consistently achieves strong retain accuracy while effectively reducing performance on the forgotten classes. These results confirm that class-level unlearning objectives are met, while maintaining state-of-the-art performance across different forgetting scenarios.

\begin{table}[htbp]
\centering
\caption{\textbf{Unlearning performance comparison for AST model on UrbanSound8K dataset across different forget set scenarios ($f=1$, $f=C/2$, and $f=C-1$)}.}
\label{tab:urban_sound_8k_accuracy}
\begin{tabular}{@{}l|cc|@{}}
\toprule
\textbf{} & \multicolumn{2}{c|}{\textbf{UrbanSound8K}} \\ \cmidrule(l){2-3}
\text{} & $Acc_r$ & $Acc_f$ \\
\midrule
$f = 1$ &\result{87.98}{00.02}  & \result{19.90}{00.21} \\
\rowcolor[HTML]{EFEFEF}
$f=C/2$ &\result{91.49}{00.03}  & \result{14.87}{00.06} \\
$f=C-1$ &\result{96.97}{00.03}  & \result{15.27}{00.02} \\
\bottomrule
\end{tabular}%
\label{tab:results_audio}
\end{table}

\subsection{Privacy Results}

To assess the privacy vulnerabilities of the trained model, we performed confidence-based Membership Inference Attack (MIA) \cite{shokri2017membership} in order to determine whether a given sample was originally part of the model's training set before class-based forgetting. This evaluation is essential for measuring privacy leaks after the forgetting step. As shown in Table~\ref{tab:MIA_table}, our method demonstrates near-perfect resistance across all datasets, indicating strong privacy guarantees. 

Finally, we assess whether removing LoRa after training allows the prompts to retain functionality, potentially enabling a jailbreak, i.e., whether an adversary could bypass the intended forgetting mechanism by reusing learned prompts. Table~\ref{tab:results_attack} evaluates this scenario. The \textit{Full Knowledge} model refers to a baseline trained with access to all classes and using both prompts and LoRa adapters, without any unlearning. In this setting, we observe that classification accuracy remains high even after the LoRa module is removed, indicating that learned knowledge has been partially absorbed by the prompts and the classification head. This suggests a vulnerability: residual information may remain accessible through prompts alone, posing a risk of unauthorized knowledge recovery.

In contrast, our method experiences a sharp accuracy drop under the same LoRa-removal condition, demonstrating that prompts by themselves are ineffective without the LoRa components. This behavior highlights a key security advantage of our design—class knowledge is tightly coupled to both the prompt tokens and their corresponding LoRa paths, and becomes inaccessible once LoRa is detached. As a result, even if prompts are exposed or reused, they do not yield meaningful predictions. This architecture reinforces our method’s robustness against jailbreak attempts and ensures stronger guarantees for secure, verifiable unlearning.

\begin{table}[ht]
\centering
\setlength{\tabcolsep}{2pt}
\caption{\textbf{Resistance to Membership Inference Attack (MIA)}. Percentage probability of successful MIA in the case $f = C{-}1$. Lower is better. }
\begin{tabular}{c|cccccc}
\toprule
\textbf{Method }& \multicolumn{2}{c}{\textbf{BloodMNIST}}                        & \multicolumn{2}{c}{\textbf{DermaMNIST}}  & \multicolumn{2}{c}{\textbf{OrganSMNIST}}\\
                     \midrule

\textbf{SSD~\cite{foster2024fast}} & \multicolumn{2}{c}{12.38}  & \multicolumn{2}{c}{36.71} & \multicolumn{2}{c}{32.98} \\
\rowcolor{gray!10}
\textbf{Goel et al.~\cite{goel2022towards}} & \multicolumn{2}{c}{00.00}  & \multicolumn{2}{c}{00.32} & \multicolumn{2}{c}{{00.12}} \\

\textbf{LMEraser~\cite{xu2025lmeraser}}  & \multicolumn{2}{c}{12.67}  & \multicolumn{2}{c}{37.85} & \multicolumn{2}{c}{10.04} \\
\rowcolor{gray!10}
\textbf{Ours} & \multicolumn{2}{c}{00.00}  & \multicolumn{2}{c}{00.12} & \multicolumn{2}{c}{00.02} \\
\bottomrule
\end{tabular}
\label{tab:MIA_table}
\end{table}

\begin{table}[ht]
\centering
\setlength{\tabcolsep}{2pt}
\caption{\textbf{Impact of LoRa removal on classification accuracy}. Our method experiences a sharp accuracy drop, ensuring robustness against jailbreak attempts, while the baseline retains performance, indicating residual knowledge.}
\begin{tabular}{c|cccccc}
\toprule
\textbf{Method }& \multicolumn{2}{c}{\textbf{BloodMNIST}}                        & \multicolumn{2}{c}{\textbf{DermaMNIST}}  & \multicolumn{2}{c}{\textbf{OrganSMNIST}}\\
                     \midrule

\textbf{Full Knowledge} & \multicolumn{2}{c}{75.65}  & \multicolumn{2}{c}{65.14} & \multicolumn{2}{c}{49.81} \\
\rowcolor{gray!10}
\textbf{Ours} & \multicolumn{2}{c}{11.74}  & \multicolumn{2}{c}{31.50} & \multicolumn{2}{c}{08.74} \\
\bottomrule
\end{tabular}
\label{tab:results_attack}
\end{table}

Importantly, while LMeraser achieves lower forget accuracy than our method—as shown in Tables~\ref{tab:results_1},~\ref{tab:results_n2}, and~\ref{tab:results_n3}—this advantage comes with critical trade-offs. In addition to requiring access to training data and full retraining of affected clusters, LMeraser exhibits serious privacy vulnerabilities. As reported in Table~\ref{tab:MIA_table}, it is significantly more susceptible to membership inference attacks (MIA), with higher success rates across all datasets. In contrast, our method achieves near-perfect resistance to MIA, preventing recovery of forgotten information even under adversarial probing. 

Cumulatively, all results highlight the robustness of our approach, which combines efficient, retraining-free unlearning with strong privacy guarantees—preventing unauthorized knowledge recovery and making it resistant to jailbreak attempts. As a result, it is well-suited for real-world, regulation-compliant deployments where security and accountability are critical.

\section{Conclusion}
In this work, we introduced a novel unlearning framework that tightly couples learning and unlearning within a single, unified training phase. By associating class-level knowledge with dedicated, learnable prompt tokens, our method enables instant and verifiable unlearning through prompt removal—requiring no retraining, no access to original data, and no modification to the backbone architecture. This stands in sharp contrast to traditional unlearning approaches, which often depend on costly post hoc procedures and are difficult to scale or deploy in real-time settings.

Through extensive experiments across diverse medical imaging tasks and an additional evaluation on audio data, we demonstrated that our method achieves near-random performance on forgotten classes while maintaining high accuracy on retained ones. It also delivers robust privacy guarantees, including resistance to membership inference attacks, without compromising computational efficiency. Compared to state-of-the-art techniques like LMeraser, our framework achieves competitive forgetting performance at a fraction of the cost and model size, and crucially, without requiring additional epochs of retraining per unlearning operation.

These results underscore the unique advantages of our method: it is modular, scalable, data-free, and training-free—qualities that are essential for AI systems operating in privacy-sensitive, continuously evolving environments. By embedding removability into the model’s architecture, we lay the foundation for a new class of ethically responsive, deployment-ready foundation models.

In conclusion, prompt-based optimization offers a powerful and interpretable mechanism for adaptive unlearning, shifting the paradigm from reactive remediation to proactive, pre-forgettable design. We believe this work marks a step forward in the development of foundation models that not only learn well, but also forget responsibly.

\begin{acks}
Rutger Hendrix (conceptualization, code writing, experiments and paper writing) acknowledges the support of the PNRR ICSC National Research Centre for High Performance Computing, Big Data and Quantum Computing (CN00000013), under the NRRP MUR program funded by NextGenerationEU. Giovanni Patanè (experiments and evaluation) has been supported by MUR under PNRR Mission 4, Component 2, Investment 1.1, PRIN project RESILIENT (CUP E53D23016360001).
Simone Carnemolla, Federica Proietto Salanitri, Giovanni Bellitto, Concetto Spampinato, and Matteo Pennisi have been supported by the European Union – Next Generation EU, Mission 4 Component 2 Line 1.3, through the PNRR MUR project PE0000013 – FAIR “Future Artificial Intelligence Research” (CUP E63C22001940006). Their contributions span conceptualization and design of the method (Spampinato, Carnemolla), implementation and code development (Proietto Salanitri, Bellitto), and paper writing and validation (Spampinato, Pennisi).
R. Hendrix and G. Patanè are PhD students in the National PhD in Artificial Intelligence Health and Life Sciences at Università Campus Bio-Medico di Roma, in the 38th cycle bis and XL cycle, respectively.
\end{acks}

\bibliographystyle{ACM-Reference-Format}
\balance
\bibliography{biblio}


\end{document}